\documentclass[conference]{IEEEtran}
\pdfoutput=1
\IEEEoverridecommandlockouts

\usepackage{cite}
\usepackage{amsmath,amssymb,amsfonts}
\usepackage{algorithmic}
\usepackage{graphicx}
\usepackage{textcomp}
\usepackage{xcolor}

\definecolor{cblue}{RGB}{31,119,180}
\definecolor{clightblue}{RGB}{127,179,213}
\definecolor{corange}{RGB}{255,127,14}
\definecolor{clightorange}{RGB}{255,187,120}
\definecolor{cgreen}{RGB}{44,160,44}
\definecolor{cred}{RGB}{214,39,40}
\definecolor{cpurple}{RGB}{148,103,189}
\definecolor{cgray}{RGB}{110,110,110}

\usepackage{booktabs}
\usepackage{multirow}
\usepackage{array}
\usepackage{tikz}
\usepackage{pgfplots}
\pgfplotsset{compat=1.18}
\usetikzlibrary{shapes.geometric, shapes.multipart, shapes.symbols, arrows.meta, positioning, fit, backgrounds, calc, matrix, decorations.pathreplacing, shadows}
\usepackage{enumitem}
\usepackage[hidelinks]{hyperref}
\usepackage{cleveref}
\usepackage{fontawesome5}

\newcommand{\KP}{\mathsf{KP}}
\newcommand{\header}{\mathsf{H}}
\newcommand{\body}{\mathsf{B}}
\newcommand{\meta}{\mathsf{M}}
\newcommand{\fitness}{\mathcal{F}}

\newcommand{\variation}{\mu}
\newcommand{\selection}{\sigma}
\newcommand{\recombination}{\rho}
\newcommand{\inheritance}{\lambda}
\newcommand{\local}{\mathsf{local}}

\begin{document}

\title{Silicon Language: A Robot-Native Knowledge Exchange Framework for Heterogeneous Robots}

\author{
    \IEEEauthorblockN{Yi Liu\textsuperscript{1,2},
        Xianglin Meng\textsuperscript{2},
        Chang Chen\textsuperscript{3},
        and Jingjing Fan\textsuperscript{3,*}}
    \IEEEauthorblockA{\textsuperscript{1}School of Mechanical Engineering,
        Beijing Institute of Technology, Beijing 100081, China}
    \IEEEauthorblockA{\textsuperscript{2}Yulin Saiyi Intelligent Technology Co., Ltd., Yulin 719000, China}
    \IEEEauthorblockA{\textsuperscript{3}Yulin Intelligent Unmanned Equipment Innovation Center Co., Ltd., Yulin 719000, China}
    \IEEEauthorblockA{\textsuperscript{*}Corresponding author: Jingjing Fan}
}

\maketitle

\begin{abstract}
    Reusing a capability across heterogeneous robots still requires substantial human adaptation and verification: transferring a skill often means re-engineering interfaces, retuning parameters, and re-validating safety. We introduce Silicon Language, a robot-native knowledge exchange framework that treats the robot as the active subject of its own capability evolution. In this framework, a robot that wants a capability encodes its own experience into knowledge packets, publishes them, retrieves peer packets, translates them for its own sensors and actuators, and reviews them through independent local trial. A receiver-side usability evaluation procedure lets each robot decide for itself whether an external packet is useful, and progressive blending with automatic rollback is designed to reduce the risk of negative transfer when adopting it. The system combines a three-layer infrastructure---edge agent, Silicon Transfer Protocol (STP), and knowledge hub---with a capability stack inspired by the human scholarly system. We report a 30-day proof-of-concept deployment at an above-ground simulated-mine laboratory in Yulin, with following trials on an outdoor sand road and an indoor factory floor. Two heterogeneous robots encoded and translated three capabilities across embodiments through operator-assisted file copies mediated by the Silicon Language translation layer; source-side trials of the dust-locked following behavior were recorded on Taurus. Project records indicate that per-capability adaptation time dropped from days to hours; we present these figures as descriptive deployment records rather than controlled measurements. The deployment provides initial evidence for cross-embodiment knowledge exchange; fleet-level autonomous evolution and controlled with/without-packet comparisons remain future work.
\end{abstract}


\section{Introduction}
\label{sec:intro}

Reusing a capability across heterogeneous robots still requires substantial human adaptation. Engineers label data, write code, design training pipelines, evaluate outcomes, and deploy updates. When robot A learns a useful skill in one environment, robot B---with different sensors, actuators, and constraints---cannot inherit it directly; humans must repeat much of the adaptation and verification cycle for B. In this prevailing workflow the robot is a knowledge \emph{object}: humans decide what to transfer, how to adapt it, and whether it is safe. The robot itself is not the agent of its own capability growth.

This human-centric bottleneck caps the scalability of robot fleets: as the number of robots, environments, and required capabilities grows, the human effort required grows with it. Human researchers, by contrast, operate inside a scholarly ecosystem: they write papers, publish them, cite prior work, translate ideas across fields, subject claims to peer review, revise versions, and debate at conferences. What is needed is not merely a faster human-driven pipeline, but a \emph{robot-native} scholarly system in which the robot itself is the active subject---a robot that wants a capability encodes its own experience, publishes it, retrieves peer knowledge, translates it for its own embodiment, reviews it through independent trial, and revises it through versioned iteration.

Prior work has addressed pieces of the problem, but usually keeps agency on the human side. RoboEarth and KnowRob let robots share semantic maps and action recipes\cite{waibel2011roboearth, tenorth2013knowrob}; RoboEarth in particular demonstrated cross-robot upload, retrieval, and reuse across heterogeneous platforms. However, the shared representations and action recipes were largely human-predesigned, and the receiving robot did not independently judge whether an incoming artifact was useful or safe for its own embodiment through a local trial protocol. RoboBrain builds large knowledge graphs for cross-modal reasoning\cite{saxena2014robobrain}, yet it does not support receiver-side adaptation, publication, retrieval, review, or iteration across robots. Federated learning and Open X-Embodiment aggregate gradients or data\cite{mcmahan2017communication, openxembodiment}, but what is shared cannot leave the training process, and robots do not decide what is worth sharing. ROS/DDS and FABRIC provide communication and coordination\cite{quigley2009ros, dds, fabric}, while IEEE 2751 standardizes 3-D map-data exchange\cite{ieee2751}. Recent agentic systems narrow the gap from another direction: Voyager lets an LLM-driven agent autonomously write, execute, and store skills as code in Minecraft\cite{wang2023voyager}; Code as Policies shows that language-model programs can express robot control logic on real robots\cite{liang2023codeaspolicies}; and embodied foundation models such as {PaLM-E}\cite{driess2023palme}, {RT-2}\cite{brohan2023rt2}, RoboCat\cite{bousmalis2023robocat}, and Octo\cite{octo2024} further scale generalizable skill generation and reuse\cite{firoozi2024foundation}. These systems demonstrate that machines can produce and reuse executable knowledge; they do not address the cross-embodiment, cross-robot exchange problem, in which the receiving machine has different sensors, actuators, and constraints and must independently judge whether externally produced knowledge is useful and safe \emph{for itself}.

We introduce \textbf{Silicon Language}, a robot-native knowledge exchange framework whose central design principle is that the robot is the subject of its own capability evolution. The framework provides the lifecycle primitives a robot needs to act as a knowledge \emph{subject}: \emph{writing} its experience into packets, \emph{publishing} them, \emph{retrieving} peer packets, \emph{translating} them to its own morphology, \emph{reviewing} them through local trial, and \emph{revising} them through versioned iteration. Our contributions are:
\begin{enumerate}[label=\textbf{C\arabic*.}, leftmargin=2.2em, itemsep=1pt]
    \item \textbf{A robot-native, robot-centric knowledge exchange framework.} A three-layer infrastructure (edge agent, Silicon Transfer Protocol, knowledge hub) plus a knowledge-packet abstraction and a receiver-side review protocol, organized as a capability stack inspired by the human scholarly system. The receiving robot---not a human engineer and not the publisher---decides what to adopt and how to validate it for its own embodiment (\cref{sec:method}).
    \item \textbf{Encoding-independent usability evaluation with negative-transfer risk mitigation.} Receiving robots independently judge whether a published packet is useful, partially useful, or unusable for them through local trial. Progressive blending and automatic rollback are designed so that the fallback outcome of a failed adoption is a return to the prior software state (\cref{subsec:usability,sec:formalization}).
    \item \textbf{A robot-centric proof-of-concept deployment.} A 30-day deployment with two heterogeneous robots at an above-ground laboratory configured to simulate an underground coal mine in Yulin. The reported Taurus following trials took place separately on an outdoor sand road and an indoor factory floor, not inside the simulated mine. The robots encoded, published, and translated capabilities across embodiments on their own edge agents. The cross-robot file copy was operator-assisted only because the production STP exchange is not yet running. Per-capability adaptation time dropped from days to hours in our engineering records (\cref{sec:experiments}). We report these figures as descriptive deployment records, not controlled measurements.
\end{enumerate}

We deliberately scope this paper as a \emph{proof of concept} for the exchange framework. The empirical study is small (two robots, one simulated mine environment, one deployment cycle), the efficiency baselines rely partly on engineering estimates, and the fitness-driven evolutionary mechanisms described in \cref{subsec:evolution} are part of the framework design but not yet validated at fleet scale; their implementation status will be confirmed in the code audit. We state these limits explicitly in \cref{sec:discussion}.

\begin{figure*}[!t]
\centering
\includegraphics[width=\linewidth]{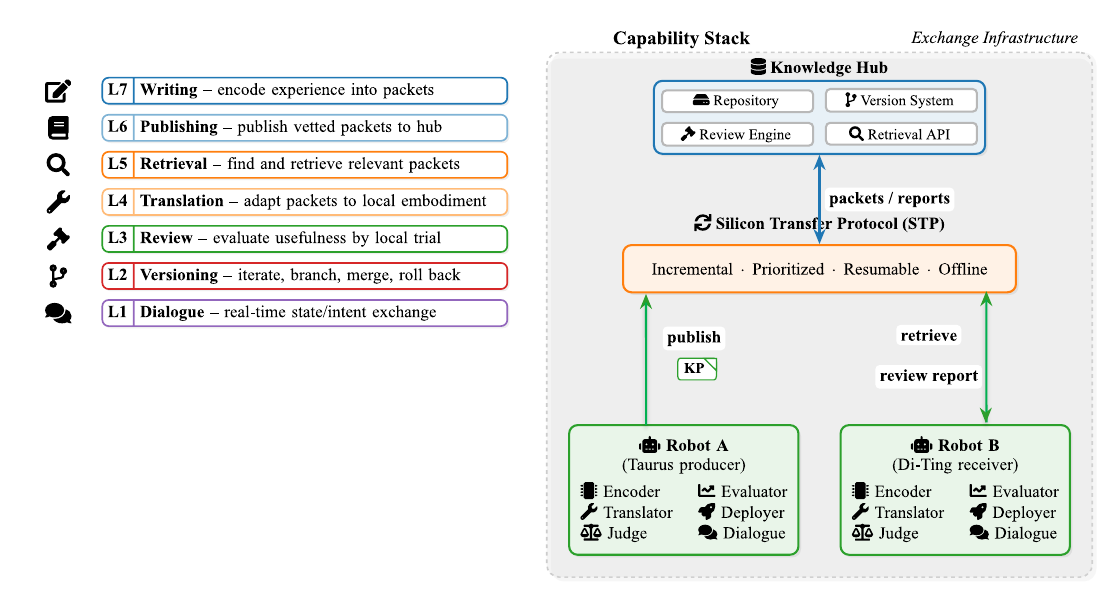}

\caption{Silicon Language reference architecture. The three-layer infrastructure (edge agent, STP, knowledge hub) supports a seven-layer capability stack inspired by the human scholarly system. The stack is a functional organization, not a claim of equivalence with human scholarship.}
\label{fig:architecture}
\end{figure*}

\section{Related Work}
\label{sec:related}

\paragraph{Robot knowledge repositories.} Early systems such as RoboEarth and KnowRob provided Web-accessible knowledge bases in which robots could share semantic maps, object models, and action recipes\cite{waibel2011roboearth, tenorth2013knowrob}. RoboEarth in particular showed that robots could upload experience, retrieve peer knowledge, and reuse it across heterogeneous platforms through a shared representation and a robot-specific skill-abstraction layer. These systems established that cross-robot knowledge reuse is possible, but the knowledge formats and task recipes were largely human-predesigned, and the receiving robot did not independently judge whether an incoming artifact was useful or safe for its own embodiment through a local trial protocol. RoboBrain later scaled cross-modal knowledge graphs for single-robot reasoning\cite{saxena2014robobrain}, yet it did not address receiver-side adaptation, publication, retrieval, review, or iterative evolution across a fleet.

\paragraph{Distributed and federated learning.} Federated learning lets multiple agents train models while keeping data local\cite{mcmahan2017communication}. However, federated learning shares gradients or weights, not reusable knowledge artifacts; aggregation rules are human-designed; and knowledge cannot leave the training run. Open X-Embodiment demonstrated the power of pooling diverse robot data for centralized training\cite{openxembodiment}, but the data are human-collected and curated, and robots remain objects of data collection rather than subjects of knowledge production.

\paragraph{Executable skills and code generation.} Voyager showed that an LLM-driven agent can autonomously write, execute, verify, and store skills as code, building a growing skill library in Minecraft\cite{wang2023voyager}. Code as Policies demonstrated that language-model programs can express reactive robot control logic, with deployments on several real robot platforms\cite{liang2023codeaspolicies}. More recently, vision-language-action models and embodied multimodal language models have transferred web-scale semantics directly into robot control\cite{brohan2023rt2,driess2023palme}; RoboCat and Octo train generalist manipulation policies across multiple embodiments\cite{bousmalis2023robocat,octo2024}; and diffusion policies provide a flexible generative model for visuomotor skills\cite{chi2023diffusion}. These works establish that machines can produce and reuse executable knowledge. The distinguishing problem we target is \emph{cross-embodiment exchange between independent robots}: the receiving machine has different sensors, actuators, and constraints from the producer, and must independently adapt the incoming knowledge and judge---through its own trial protocol---whether it is useful and safe to adopt. Neither a shared skill library nor human-prompted code generation addresses this receiver-side adaptation and acceptance problem.

\paragraph{Robot communication and coordination.} ROS topics and DDS provide real-time message passing\cite{quigley2009ros, dds}; FABRIC adds decentralized coordination with identity verification and encrypted data sharing\cite{fabric}. IEEE 2751 standardizes 3-D map-data exchange\cite{ieee2751}. These systems enable the \emph{dialogue} layer of robot interaction, but they do not provide the higher layers of the exchange lifecycle: writing, publishing, retrieving, translating, reviewing, and versioning knowledge.

\paragraph{Knowledge transfer and lifelong learning.} Learngene and related work study how neural-network knowledge can be decomposed, compressed, and transferred across tasks\cite{wang2022learngene}. Progressive neural networks and related methods accumulate task-specific columns\cite{rusu2016progressive}. Recent benchmarks such as {LIBERO} formalize forward transfer and negative backward transfer in lifelong robot learning\cite{liu2023libero}, and Reset \& Distill directly targets negative transfer in continual reinforcement learning\cite{ahn2024reset}. Cross-embodiment methods align latent state-action spaces\cite{wang2024crossembodiment}, discover shared skills from human and robot videos\cite{xu2023xskill}, encode embodiment descriptors via transformers\cite{sferrazza2024body}, or train one policy across diverse locomotion morphologies\cite{bohlinger2024onepolicy}. These approaches address the technical question of \emph{how} to transfer knowledge, but not the ecological question of \emph{who} decides what to publish, \emph{who} validates it, and \emph{how} it evolves over time.

\paragraph{Positioning.} Silicon Language is distinguished by two properties. First, it treats knowledge as a first-class, versioned, reviewable artifact (a knowledge packet) rather than a by-product of training. Second, the receiving robot---not the publisher, not a human---decides through a defined trial protocol whether a packet is useful for its own embodiment, which is the property that makes cross-embodiment exchange viable without human re-engineering.

\section{The Silicon Language Framework}
\label{sec:method}

Silicon Language comprises a three-layer infrastructure and a seven-layer capability stack (\cref{fig:architecture}). The infrastructure layers are the \emph{edge agent} (resident on each robot), the \emph{Silicon Transfer Protocol} (STP), and the \emph{knowledge hub}. The capability stack is a functional organization inspired by the human scholarly system: writing (L7), publishing (L6), retrieval (L5), translation (L4), review (L3), versioning (L2), and dialogue (L1). We use the scholarly terms as a naming and design guide; we do not claim functional equivalence with human scholarship. The overarching design goal is that the robot, not a human engineer, is the active subject of the exchange: it decides what experience to encode, what packets to retrieve, and whether an incoming packet is useful and safe for its own embodiment.

\subsection{Three-layer infrastructure}
\label{subsec:infrastructure}

\textbf{Edge agent.} Every robot joining the system runs an edge agent with six modules, so that the robot itself can act as the subject of knowledge exchange. The \emph{knowledge encoder} monitors the robot's own execution and writes its experience into knowledge packets. The \emph{effect evaluator} measures local performance before the robot publishes a packet and after the robot adopts one. The \emph{translation engine} adapts foreign packets to the robot's own sensors, actuators, and compute. The \emph{redeployer} installs adapted packets with hot-swap and rollback support. The \emph{usability judge} runs the four-step evaluation in \cref{subsec:usability} and renders the robot's own verdict. The \emph{dialogue module} exchanges parametric state and intent vectors with nearby robots.

\textbf{Silicon Transfer Protocol.} STP is an application-layer protocol above TCP/IP and DDS, optimized for knowledge-packet transport. It follows four principles: incremental updates (send differences, not whole packets), priority scheduling (safety-critical before task-critical before background), resumable transfer, and compressed representation. STP defines ten message types; see \cref{tab:stp}. In the deployment of \cref{sec:experiments}, STP is specified but not yet running in production; the transfers were executed as operator-assisted file copies through the Silicon Language translation layer.

\textbf{Knowledge hub.} The hub is logically centralized but physically distributed. It contains a repository, a version-management system, a review/fitness engine, and a retrieval API. It supports three deployment modes: edge servers (for subterranean facilities and other closed environments), cloud, and fully decentralized peer-to-peer gossip. In the deployment of \cref{sec:experiments}, the hub ran on a single edge server in the laboratory; the gossip mode was not exercised in the simulated mine.

\subsection{Seven-layer capability stack}
\label{subsec:sevenlayers}

\cref{tab:layers} maps each layer of the human scholarly system to its Silicon Language counterpart. The mapping is a design guide: each layer has concrete protocol primitives and algorithmic operations, but we make no claim that, e.g., machine review reproduces the epistemic function of human peer review.

\begin{table}[t]
\centering
\caption{Mapping between the human scholarly system and the Silicon Language capability stack.}
\label{tab:layers}
\begin{tabular}{@{}lll@{}}
\toprule
Human system & Function & Silicon Language layer \\
\midrule
Writing & Solidify experience & L7 Writing \\
Publishing & Enter shared repository & L6 Publishing \\
Citation / search & Build on prior work & L5 Retrieval \\
Translation & Cross-domain reuse & L4 Translation \\
Peer review & Validate claims & L3 Review \\
Versioning & Continuous revision & L2 Versioning \\
Dialogue & Real-time exchange & L1 Dialogue \\
\bottomrule
\end{tabular}
\end{table}

\subsection{Knowledge packets and machine-native knowledge forms}
\label{subsec:knowledgepacket}

The atomic unit of Silicon Language is the \emph{knowledge packet} $\KP=(\header,\body,\meta)$. The header $\header$ contains a producer identifier, knowledge type, semantic version, current fitness score, applicability description, dependency list, and access-control policy. The body $\body$ carries the knowledge content itself. The metadata $\meta$ records training episodes, an environment fingerprint, a hardware profile, and the performance delta relative to the previous version.

\begin{figure*}[t]
\centering
\includegraphics[width=\textwidth]{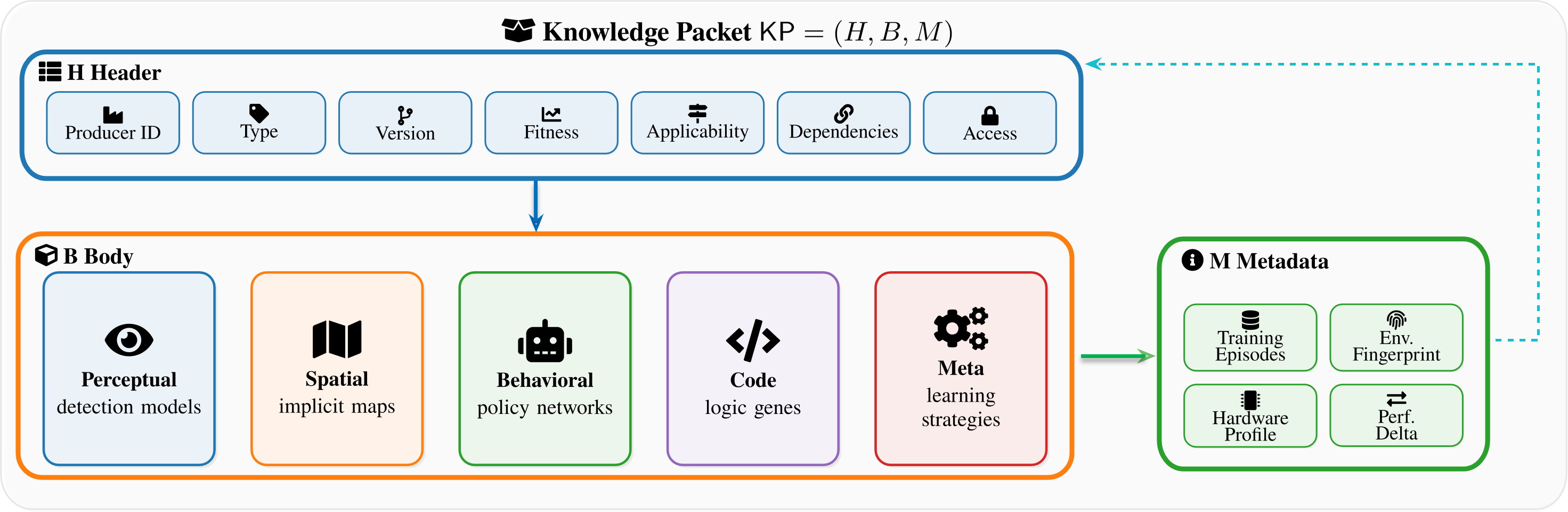}
\caption{Structure of a Silicon Language knowledge packet. The header specifies identity and applicability; the body supports five knowledge forms; metadata records training, environment, hardware, and performance information.}
\label{fig:packet}
\end{figure*}

The body can take five machine-native forms, summarized in Fig.~\ref{fig:packet}. \emph{Perceptual knowledge} is parameter models for detection and scene understanding. \emph{Spatial knowledge} includes neural implicit maps, topological maps, and traversability models. \emph{Behavioral knowledge} comprises policy networks and motion skills. \emph{Code knowledge} represents processing pipelines as directed graphs whose nodes encode a structured task description (the exact artifact format is under audit; it may include code, configuration, and natural-language constraints). The receiving robot generates or retrieves an implementation for its own hardware. \emph{Meta-knowledge} captures learning strategies, curricula, hyperparameters, and adaptation guides. The set of forms is open: the framework standardizes the \emph{interface} (header fields, body form tag, trial protocol, and review report format), not the encoding.

\paragraph{A concrete packet.} \cref{tab:packetexample} instantiates this schema for the dust-locked lead-vehicle-following packet transferred in \cref{sec:experiments}. The packet identifier and file paths are recorded in the data package so that the example can be traced to raw source and bag records.

\begin{table*}[t]
\centering
\caption{Example knowledge packet for dust-locked lead-vehicle following, produced on Taurus and adapted for Di-Ting. The body represents processing logic, while implementation code and message bindings depend on the robot embodiment.}
\label{tab:packetexample}
\begin{tabular}{@{}p{3.2cm}p{13.5cm}@{}}
\toprule
Field & Content \\
\midrule
Packet ID & \texttt{kp.behavior.following.dust\_locked.v1} \\
Producer & Taurus (\texttt{guide\_vehicle\_tracking} package, Autoware-era branch, $\sim$2026-Q2) \\
Type & Code knowledge (processing-logic gene: 9-stage pipeline + state machine) \\
Version & 1.0.0 (supersedes: none) \\
Body (pipeline) & S1 ROI crop $\rightarrow$ S2 ground removal $\rightarrow$ S3 voxel downsample $\rightarrow$ S4 Euclidean cluster $\rightarrow$ S5 OBB fit $\rightarrow$ S6 target selection $\rightarrow$ S7 state machine (IDLE/\allowbreak SEARCHING/\allowbreak LOCKED/\allowbreak REACQUIRE/\allowbreak SAFE\_STOP) $\rightarrow$ S8 distance-to-speed lookup $\rightarrow$ S9 lateral controller (Ackermann $\delta$ or differential $\omega$) \\
Body (key knobs) & Distance-to-speed breakpoints $\{0,2,3.5,5,7,9,12,15\}$\,m $\mapsto$ speeds $\{0,0,1.0,1.6,2.5,4.0,6.0,6.0\}$\,m/s; stop distance 5\,m; miss threshold 20 frames; reacquire timeout 3\,s \\
Applicability & Non-structured road; single lead vehicle (car/truck/bus); visibility below lidar range; optional radar backup for dust \\
Dependencies & LiDAR point cloud $\ge$16 channels, $\ge$10\,Hz; odometry; chassis report; behavior command \\
Safety rules & Emergency stop overrides all commands; SAFE\_STOP after lost-lock timeout; minimum stop-distance constraint (structural) \\
Safety parameters & Recomputed per chassis: max speed (1.0\,m/s Di-Ting, 5.56\,m/s Taurus), stop distance (5\,m on Taurus source), lateral gain, steering/yaw-rate limits \\
Metadata & Source: Taurus 5 trials, 1,037\,s, 771.9\,m aggregated motion; trial~4 operator-labelled failed; lock-retention/safe-distance metrics not recorded \\
\bottomrule
\end{tabular}
\end{table*}

The packet design separates what is exchanged verbatim from what must be grounded in the receiver. Safety rules, such as the priority of an emergency-stop command, are structural invariants and travel unchanged. Safety parameters, such as maximum speed, stop distance, or controller gains, are recomputed by the translation engine under the constraint that they be at least as conservative as the publisher's, and are then revalidated during the trial protocol of \cref{subsec:usability}. This split mirrors the broader safe reinforcement-learning distinction between task objectives and safety constraints\cite{gu2024safe}. This interface-level standardization lets robots exchange knowledge regardless of whether a body is stored as network weights, a map file, or a code graph.

\subsection{Encoding-independent usability evaluation}
\label{subsec:usability}

\begin{figure}[t]
\centering
\resizebox{0.95\linewidth}{!}{%
\begin{tikzpicture}[
    node distance=0.4cm,
    stepbox/.style={rectangle, rounded corners, draw, minimum width=2.0cm, minimum height=0.8cm, align=center, font=\small},
    arrow/.style={-{Stealth[length=2mm]}, thick}
]
\node[stepbox, fill=yellow!15] (s1) {Pre-screening};
\node[stepbox, right=of s1, fill=yellow!25] (s2) {Translation};
\node[stepbox, right=of s2, fill=yellow!35] (s3) {Trial};
\node[stepbox, right=of s3, fill=yellow!45] (s4) {Verdict};

\draw[arrow] (s1) -- (s2);
\draw[arrow] (s2) -- (s3);
\draw[arrow] (s3) -- (s4);

\node[stepbox, below=0.6cm of s3, fill=red!10] (rb) {Rollback};
\draw[arrow, dashed] (s4) |- (rb);
\draw[arrow, dashed] (rb) -| (s2);
\end{tikzpicture}%
}
\caption{Receiver-side usability evaluation. The receiver pre-screens, translates, trials, and verdicts each packet, with progressive blending and automatic rollback mitigating negative-transfer risk.}
\label{fig:usability}
\end{figure}
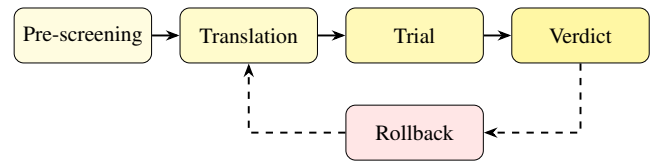

When a robot receives a knowledge packet, it does not trust the publisher's claim. Instead, it runs a four-step evaluation (\cref{fig:usability}). A clarification on scope: the evaluation \emph{protocol} is encoding-independent---pre-screening, trial, and verdict are defined over the packet interface regardless of body form---but the \emph{translation adapters} are not. Loading network weights, retargeting a map, and reconstructing a structured code representation require different adapters, each specialized to one body form.

\textbf{Step 1: applicability pre-screening.} The receiver compares the packet's applicability description, environment fingerprint, and hardware profile against its own situation. Mismatches filter out obviously unsuitable packets without costly trial.

\textbf{Step 2: translation.} The translation engine adapts the packet to the receiver's embodiment using the adapter matching the body form: direct parameter loading, interface retargeting, lightweight adapter training, or semantic-graph reconstruction. Safety rules are preserved verbatim; safety parameters are recomputed as described above.

\textbf{Step 3: trial.} The adapted packet is first tested in a digital twin or simulator. If it passes, the robot runs a limited real-world trial and records the change in task completion rate, efficiency, safety metrics, and resource consumption. Trial budgets (number of episodes, duration, environment conditions) and acceptance criteria are defined per knowledge type before the trial begins; the reported metrics are retrospective descriptions from the deployment logs.

\textbf{Step 4: verdict.} The packet is labeled \emph{usable}, \emph{partially usable}, or \emph{unusable} against the predefined acceptance criteria. Safety is evaluated as a separate gate, not as one weighted term among others: a packet that fails its safety criteria is unusable regardless of its efficiency or completion-rate scores. Partial usability means only verified fragments are retained. A negative verdict is reported to the hub as a review record.

\textbf{Negative-transfer risk mitigation.} The receiver deploys an adopted packet through progressive blending with its existing capability. For continuously parameterizable components whose structures are compatible (e.g., same-architecture network weights, controller gains), blending interpolates with a mixing coefficient $\alpha\in[0,1]$; for discrete structures (code graphs) or incompatible parameterizations, ``blending'' is realized as staged rollout---shadow mode, then limited-scope activation, then full activation. These two mechanisms play analogous roles in limiting exposure, but we do not claim they share the same performance properties. If monitored performance violates the predefined degradation criteria, the redeployer automatically restores the prior version. This mechanism is designed so that the \emph{software-state} fallback of a failed adoption is the prior state; it cannot undo time or energy spent during the trial itself, and it does not cover physical consequences of trial execution, which is why trials are first run in simulation and then in limited scope. The operative acceptance rule is concrete rather than analytic: \emph{a packet is adopted only if it meets the predefined task thresholds and passes the separate safety gate; rollback triggers when the monitored degradation criteria are met.} The observed source-side safety-stop behavior and the deployment limits are reported in \cref{sec:experiments}; adoption rollback was not exercised in the recorded trials.

\subsection{Cluster review and the fitness loop}
\label{subsec:evolution}

A packet is not adopted by the fleet just because it was published. Each receiving robot independently retrieves it, evaluates it in its own environment, and uploads a review report containing environment descriptions, performance deltas, applicability notes, and execution logs. The knowledge hub aggregates these reports into a \emph{fitness score} (\cref{sec:formalization}). High-fitness packets receive a recommendation flag and rank higher in retrieval; low-fitness packets are marked for improvement or retired. Fitness also decays over time if a packet is neither retrieved nor positively reviewed, modeling knowledge forgetting.

These mechanisms give rise to a fleet-level loop: \emph{variation} (a robot incorporates local experience into a new version), \emph{selection} (fitness-driven retention), \emph{recombination} (merging independently improved branches), and \emph{inheritance} (validated improvements are republished and adopted by others). In our deployment, versioning and single-recipient review were exercised end-to-end; one source-side SAFE\_STOP was observed on Di-Ting during a downhill human-track failure, and one additional safety-stop path is present in the dust-locked source code but was not observed in the recorded bags (\cref{tab:rollback}). Multi-reviewer aggregation, branch recombination, and long-horizon fitness dynamics were \emph{not} exercised at fleet scale (two robots, one deployment cycle); we therefore present the loop as a designed mechanism and leave its fleet-level validation to future work (\cref{sec:discussion}).

\section{Formalization and Protocol}
\label{sec:formalization}

\subsection{Formal model}
\label{subsec:formalmodel}

A knowledge packet is a tuple $\KP_i^v=(\header_i, \body_i, \meta_i)$, where $i$ is a unique identifier and $v$ is a semantic version. The header $\header_i$ is a tuple of fields defined over finite alphabets; the body $\body_i$ is an encoding-agnostic knowledge representation at the interface level; and $\meta_i$ records provenance and performance data.

\paragraph{Fitness from multiple reviews.} Let $R_i$ be the set of robots that have submitted review reports for $\KP_i$. Each report $r\in R_i$ contains normalized deltas $\Delta_k^{(r)}\in[-1,1]$ for $k\in\{c,e,r\}$, denoting \emph{improvement} over the reviewer's pre-trial baseline in task completion rate, efficiency, and resource consumption respectively---all three are signed so that positive values mean better performance (for resource consumption, $\Delta_r$ is the \emph{reduction} in consumption, so an increase in consumption yields $\Delta_r<0$). Normalization is relative to the reviewer's own baseline measurement under its trial conditions; cross-robot comparability of normalized scores is an open question we flag in \cref{sec:discussion}. The fitness aggregates reviewers and metrics:
\begin{equation}
    \fitness(\KP_i) = \frac{1}{|R_i|}\sum_{r\in R_i}\;\sum_{k\in\{c,e,r\}} w_k\,\Delta_k^{(r)},
    \label{eq:fitness}
\end{equation}
where $w_k \ge 0$, $\sum_k w_k=1$, and weights are set by fleet policy (in our deployment they were chosen heuristically). When $|R_i|=0$---a packet with no reviews---$\fitness(\KP_i)$ is undefined and the packet carries an explicit \emph{unscored} state rather than a default value, so unscored packets cannot be confused with high- or low-fitness ones. Safety verdicts act as a \emph{separate gate} before fitness is even computed (\cref{subsec:usability}): a packet failing its safety criteria is retired regardless of $\fitness$, so efficiency gains cannot offset safety losses.

\paragraph{Decay.} Fitness decays when a packet is neither retrieved nor positively reviewed. Let $\tau$ be a forgetting horizon and $d(t)\in[0,1]$ a monotonic decay factor. The time-dependent fitness is
\begin{equation}
    \fitness_t(\KP_i) = \min\bigl(\fitness_0(\KP_i),\; \fitness_0(\KP_i)\, d(t)\bigr) \;\; \text{for } \fitness_0 \ge 0,
    \label{eq:decay}
\end{equation}
and $\fitness_t(\KP_i)=\fitness_0(\KP_i)$ for $\fitness_0<0$. Decay applies only to non-negative fitness: multiplying a negative score by $d(t)<1$ would move it toward zero and spuriously improve its ranking, so negative-fitness packets are excluded from decay and remain flagged. A new retrieval or positive report resets the decay timer.

\paragraph{Evolutionary operators.} The fleet-level loop is described by:
\begin{align}
    \variation &: (\KP_i^v, E_{\local}) \mapsto \KP_i^{v+1}, \\
    \selection &: \{\fitness_j\} \mapsto \{\text{keep}, \text{recommend}, \text{retire}\}, \\
    \recombination &: (\KP_i^{v_a}, \KP_i^{v_b}) \mapsto \KP_i^{v+1}, \\
    \inheritance &: \text{validated}(\KP) \mapsto \text{publish}(\KP).
\end{align}
These operators are defined for the hub design; as noted in \cref{subsec:evolution}, only variation and single-reviewer selection were exercised in our deployment.

\paragraph{Negative-transfer risk mitigation (operative rule).} Rather than an analytic guarantee, the framework specifies a concrete acceptance rule (\cref{subsec:usability}): a packet is adopted only if it meets the predefined task thresholds \emph{and} passes the separate safety gate; rollback triggers when the monitored degradation criteria are met. In our deployment, one source-side SAFE\_STOP was observed on Di-Ting during a downhill human-track failure, and one dust-locked safety-stop path is present in the source code but was not observed in the recorded bags (\cref{tab:rollback}). A formal analysis of this rule's properties (e.g., under what monitoring assumptions post-adoption performance stays within a bound of the pre-trial baseline) is future work.

\subsection{Silicon Transfer Protocol}
\label{subsec:stp}

STP is specified as an application-layer protocol above TCP/IP and DDS. Its ten message types are listed in \cref{tab:stp}; each maps to one or more layers of the capability stack. STP is design-only in the current deployment: the on-disk packet schemas under \texttt{knowledge\_packets/} are exactly the payload a GENE\_STREAM message would carry, and the per-trial \texttt{fitness\_report} blocks are exactly the payload a FITNESS\_REPORT message would carry, but no production STP exchange was running. The transfers reported in \cref{sec:experiments} were executed as operator-assisted file copies through the Silicon Language translation layer; a real two-robot STP exchange is scheduled as follow-up work.

\begin{table}[t]
\centering
\caption{STP message types and their corresponding capability-stack layers. The protocol is specified but not yet deployed in production; the table describes the reference design.}
\label{tab:stp}
\resizebox{\linewidth}{!}{%
\begin{tabular}{@{}lll@{}}
\toprule
Message & Role & Layer \\
\midrule
KNOWLEDGE\_ANNOUNCE & Advertise a new packet & L6 Publishing \\
KNOWLEDGE\_REQUEST & Query the hub for packets & L5 Retrieval \\
KNOWLEDGE\_OFFER & Propose a packet in response & L5 Retrieval \\
GENE\_STREAM & Fragmented packet transfer & Transport \\
ACKNOWLEDGMENT & Receipt and agreement & L1 Dialogue \\
FITNESS\_REPORT & Local review verdict & L3 Review \\
REVIEW\_RESULT & Structured review evidence & L3 Review \\
TRANSLATE\_REQUEST & Request morphology-specific tailor & L4 Translation \\
LIVE\_STATE & Current robot state vector & L1 Dialogue \\
LIVE\_INTENT & Intended action vector & L1 Dialogue \\
\bottomrule
\end{tabular}%
}
\end{table}

STP supports three topologies. \emph{Proximity peer-to-peer} connects robots in the same working area (latency target sub-10\,ms). \emph{Relay transport} forwards packets across working areas through intermediate nodes. \emph{Offline sync} stores packets locally when the network is unavailable and synchronizes when connectivity returns. The latter is essential for deployment in mines, where continuous connectivity cannot be assumed. All three topologies are specified but were not exercised as runtime protocol exchanges in the deployment of \cref{sec:experiments}.

\section{Deployment at a Simulated-Mine Laboratory and Auxiliary Trial Sites}
\label{sec:experiments}

\subsection{Scenario, platforms, and human involvement}

We deployed Silicon Language for 30 days at an embodied-intelligence laboratory in Yulin, Shaanxi Province, China. This above-ground facility simulates underground coal-mine conditions, including loss of satellite signal, darkness, humidity, dust, and narrow passages. Both robots operated in the simulated-mine area, but stable following was not achieved there with the recorded dust-locked behavior. The five Taurus following recordings analyzed below instead came from two distinct trial sites: four on an outdoor sand road (including one stationary pre-test) and one on an indoor factory floor. Thus the reported following metrics are not measurements from the simulated-mine area. The edge agents on the two robots performed encoding, translation, and pre-screening, while the cross-robot file copy was operator-assisted because STP is not yet running in production.

\begin{figure}[t]
\centering
\includegraphics[width=\linewidth]{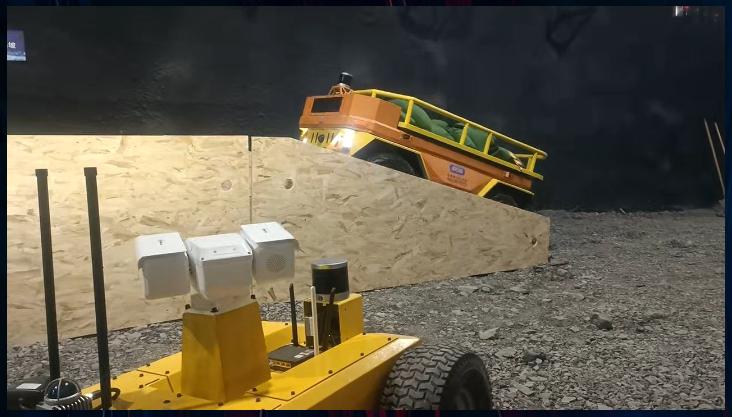}\\[4pt]
\begin{minipage}[b]{0.48\linewidth}
\centering
\includegraphics[width=\linewidth]{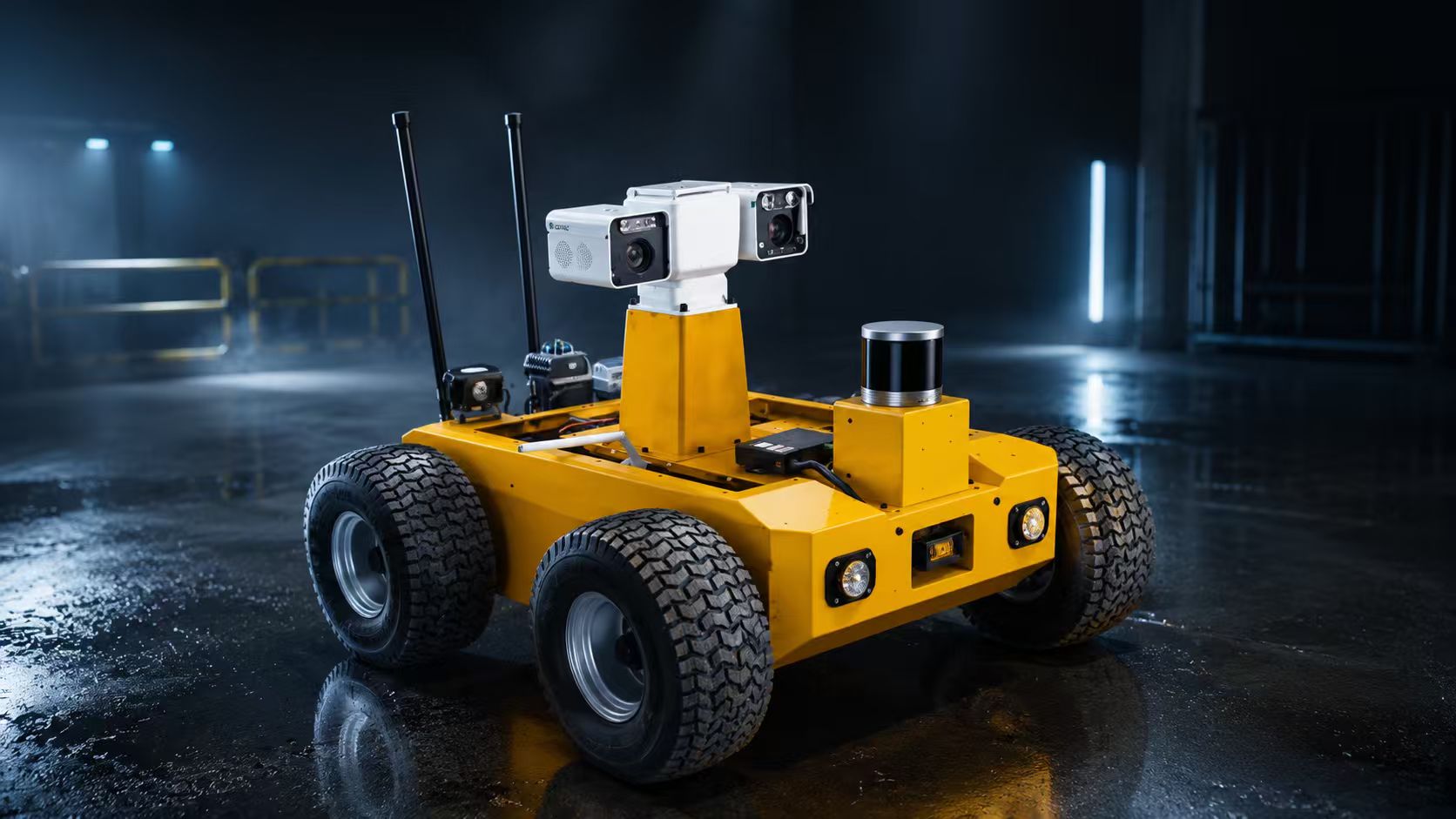}
\end{minipage}
\hfill
\begin{minipage}[b]{0.48\linewidth}
\centering
\includegraphics[width=\linewidth]{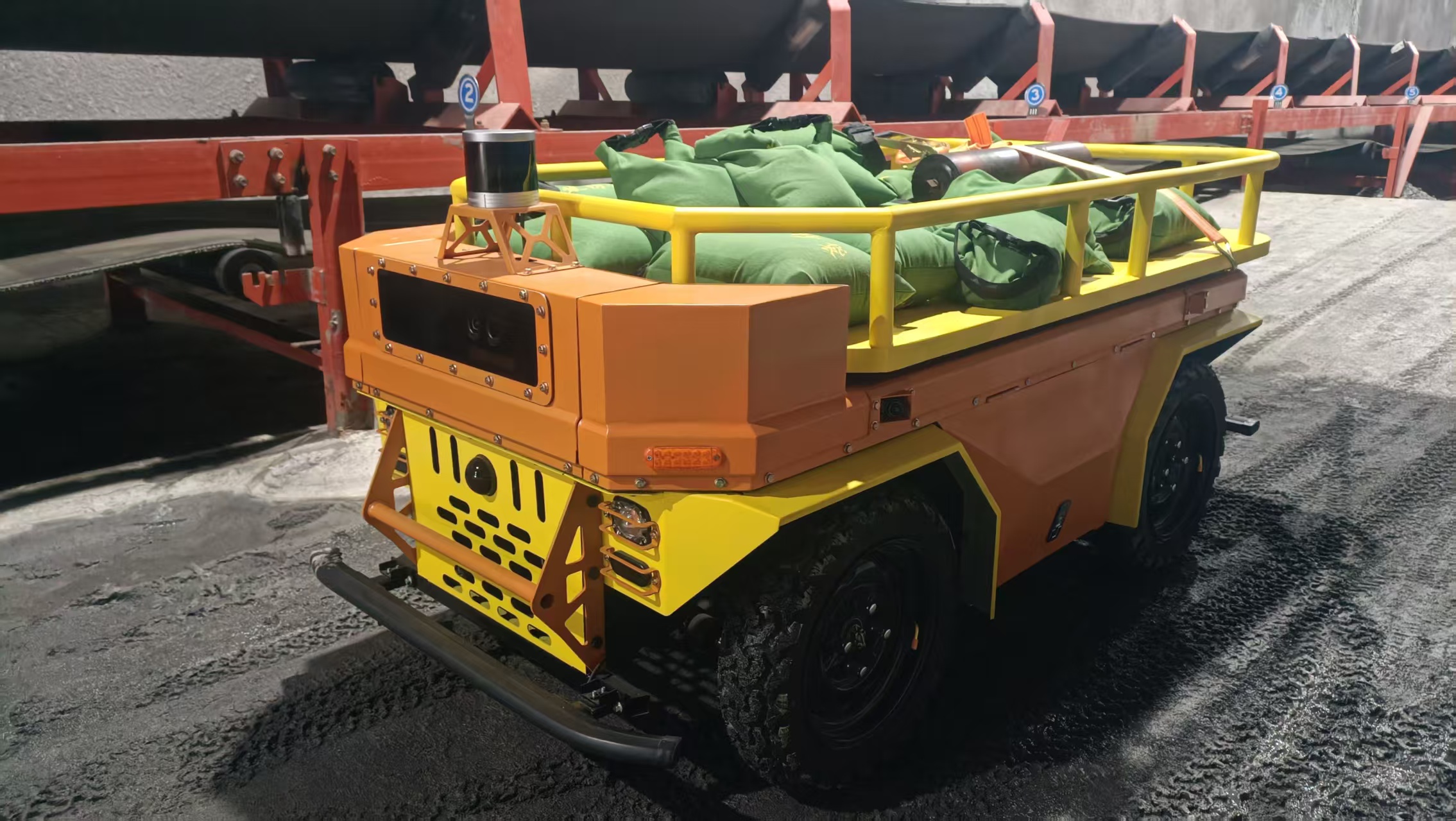}
\end{minipage}
\caption{The two robots that exchanged knowledge through Silicon Language during the deployment. \textbf{Top:} both platforms in the mine test area---foreground Di-Ting (belt inspection robot) and background Taurus (transport robot) on a test ramp. \textbf{Bottom left:} front view of Di-Ting, showing the camera/lidar mast, front lights, and communication antennas. \textbf{Bottom right:} Taurus, showing the larger transport deck, rail guides, and roof-mounted lidar/radar. The size, sensor mounting, and locomotion differences are the embodiment gaps the framework translates across.}
\label{fig:robots}
\end{figure}

\textbf{Di-Ting} (Fig.~\ref{fig:robots}, bottom left) is the belt/tunnel inspection robot: a differential-drive, 200\,kg platform measuring $0.94 \times 0.83 \times 0.84$\,m. Its sensor mast carries an RS-LiDAR-16/32 forward lidar, a rearward Livox MID360, RGB/thermal cameras, a four-gas sensor (SGA-UGT-4R), and dual communication antennas. Its regular duty is to patrol a 10\,km belt conveyor with roughly 30\,000 rollers, collecting visual, thermal, and gas data along the route.

\textbf{Taurus} (Fig.~\ref{fig:robots}, bottom right) is the material/personnel transport robot: an Ackermann-steered, 500\,kg flatbed measuring $2.01 \times 1.03 \times 1.00$\,m with a 1.09\,m wheelbase. It is equipped with three lidars (RS-LiDAR-32 forward, RS-LiDAR-Airy forward-side, RS-LiDAR-BP rearward), an LPMS-IG1 IMU, a chassis-mounted RGB camera, and a ra223f 4-D mmWave radar that is hardware-installed but whose integration was still in progress at the time of the recorded trials (its topics do not appear in the bags used here). It travels the same tunnels but at higher speed, with much larger inertia and a different braking envelope.

The two embodiments therefore differ in size, sensor mounting height, locomotion interface (differential vs.\ Ackermann), and dynamics; these are exactly the embodiment gaps that make direct code reuse unsafe and that the framework translates across. Their relevant hardware parameters are summarized in \cref{tab:platforms}. Both robots ran the full Silicon Language edge agent on an NVIDIA Jetson AGX Orin edge controller (250+\,TOPS). A small edge server in the laboratory acted as the knowledge hub. STP is specified but was not yet running in production; the transfers reported here were executed as operator-assisted file copies through the Silicon Language translation layer, with the edge agents encoding, decoding, and translating packets on each robot. \Cref{fig:robots} (top) shows the two platforms together in the laboratory test area.

\begin{table*}[!t]
\centering
\caption{Platform differences relevant to cross-embodiment transfer. Values are from chassis configuration files and author-confirmed hardware facts.}
\label{tab:platforms}
\begin{tabular}{@{}>{\raggedright\arraybackslash}p{3.6cm}>{\raggedright\arraybackslash}p{6.6cm}>{\raggedright\arraybackslash}p{6.6cm}@{}}
\toprule
 & Di-Ting (inspection) & Taurus (transport) \\
\midrule
Primary role & Belt/tunnel inspection & Material transport \\
Chassis type & Differential drive & Ackermann steering \\
Dimensions (m) & $0.94 \times 0.83 \times 0.84$ & $2.01 \times 1.03 \times 1.00$ \\
Wheelbase/track (m) & track 0.66 & wheelbase 1.09 \\
Mass & 200\,kg & 500\,kg \\
Payload & 50\,kg & 300\,kg \\
Max speed & 1.0\,m/s & 9.72\,m/s (35\,km/h) \\
Max acceleration & 0.5\,m/s$^2$ & 1.0\,m/s$^2$ \\
Endurance & 6--8\,h & 6--8\,h \\
Operating temp. & $-25$ to $60^\circ$C & $-25$ to $60^\circ$C \\
Forward lidar & RS-LiDAR-16/32 & RS-LiDAR-32 \\
Additional lidars & Livox MID360 (rearward) & RS-LiDAR-Airy (side), RS-LiDAR-BP (rearward) \\
Camera & RGB/thermal (topic \texttt{/image\_raw}) & RGB (topic \texttt{/image\_raw}) \\
mmWave radar & None & ra223f 4-D (installed, integration in progress) \\
IMU & LPMS-IG1 & LPMS-IG1 \\
Gas sensor & SGA-UGT-4R (4-gas) & None \\
Compute & NVIDIA Jetson AGX Orin, 250+\,TOPS & NVIDIA Jetson AGX Orin, 250+\,TOPS \\
Comms & Gigabit Ethernet + explosion-proof Type-C & Gigabit Ethernet + explosion-proof Type-C \\
\bottomrule
\end{tabular}
\end{table*}

\textbf{Boundary of human involvement.} The deployment was not fully autonomous, but the robot was the active subject of every step it could control. Each robot's edge agent encoded its own experience and translated incoming packets for its own embodiment; source-side trials of the dust-locked behavior were recorded on Taurus. Humans: (i) set the three capability goals and acceptance criteria; (ii) configured the edge agents and safety-rule templates before deployment; (iii) supervised real-world trials and held physical e-stop authority at all times, as required by the test-site safety protocol; (iv) executed the cross-robot packet transfers as operator-assisted file copies through the Silicon Language translation layer, because STP is not yet running in production; (v) reviewed hub logs during weekly maintenance. We report this boundary explicitly because it matters for interpreting both the efficiency and the autonomy claims below: the framework is designed for robot-driven exchange, and the operator-assisted copy step is a temporary scaffolding, not the intended normal mode.

\subsection{Transferred capabilities and the main case study}

Three capabilities were exchanged across the two embodiments:
\begin{itemize}[leftmargin=1.6em, itemsep=1pt]
    \item \emph{Intelligent following} (personnel tracking), produced on Di-Ting and translated to Taurus.
    \item \emph{Dust-locked high-speed following} (lead-vehicle tracking), produced on Taurus and translated to Di-Ting.
    \item \emph{Slope and negative-obstacle awareness}, captured as a negative gene from a Di-Ting downhill failure and used to warn Taurus away from the same failure mode.
\end{itemize}

We take dust-locked high-speed following as the main case study because it has the richest recorded evidence. The source packet is \texttt{kp.behavior.following.dust\_locked.v1}, produced on Taurus from the \texttt{guide\_vehicle\_tracking} package. Its body is a nine-stage processing logic gene: ROI crop, ground removal, voxel downsample, Euclidean clustering, oriented-bounding-box fit, target selection, a Simulink-ERT state machine (IDLE/\allowbreak SEARCHING/\allowbreak LOCKED/\allowbreak REACQUIRE/\allowbreak SAFE\_STOP), a distance-indexed velocity lookup table, and a lateral controller that emits either Ackermann steering angle or differential yaw rate depending on the receiver's chassis type. Safety rules (e.g., ``a minimum stop-distance constraint has priority over velocity commands'' and ``SAFE\_STOP after 3\,s of lost lock'') are transferred verbatim; safety parameters such as maximum speed, the numerical stop distance, and controller gains are recomputed per chassis. \Cref{tab:packetexample} summarizes the packet.

The translation to Di-Ting replaced the Ackermann lateral-controller formula with a differential yaw-rate output and loaded Di-Ting's wheel-track and speed limits. The 4-D mmWave radar was listed as an optional dust-penetrating backup in the packet header, but the recorded Taurus trials used lidar and RGB camera only; the radar integration was still in progress and no radar data appear in the bags.

\begin{enumerate}[leftmargin=1.6em, itemsep=1pt]
    \item \textbf{Origin.} The behavior was developed and first validated on Taurus in sand-road and factory-floor trials (\cref{tab:trials}).
    \item \textbf{Publication.} The edge agent encoded the invariant processing logic as a code-knowledge packet and published v1.0 to the hub.
    \item \textbf{Why direct reuse fails on Di-Ting.} Di-Ting is differential-drive, an order of magnitude smaller, and its localization point is the chassis geometric center rather than Taurus's rear axle; Taurus's steering-angle commands and clearance parameters are meaningless and unsafe on Di-Ting.
    \item \textbf{Translation.} Di-Ting's edge agent reconstructed the pipeline from the structured representation. The lidar-lock and tracking nodes were retargeted to Di-Ting's \texttt{/rslidar\_points} and \texttt{/livox/lidar} topics, and the lateral controller was re-instantiated to emit differential yaw rate with safety parameters recomputed for Di-Ting's mass and speed envelope (safety \emph{rules} retained verbatim).
    \item \textbf{Receiver-side trial.} Pre-screening and translation were completed on Di-Ting; a preliminary real-world trial was also conducted, but its detailed logs were not available for inclusion in this revision. The descriptive metrics reported here therefore come from the source side. On Taurus, the source behavior ran five recorded trials totaling 1,037\,s and 116,003 messages (\cref{tab:trials}).
    \item \textbf{Verdict and adoption.} No Di-Ting verdict is reported here because the receiver-side logs are not available. On Taurus, the source behavior ran over 771.9\,m of recorded motion; one trial was explicitly labelled as failed by the operator (trial~4), while the remaining running trials were not individually labelled (see \cref{tab:trials}). We therefore report per-trial descriptive metrics without aggregating them into a success rate.
\end{enumerate}

\subsection{Recorded following trials on Taurus}

\begin{figure*}[!t]
\centering
\includegraphics[width=\linewidth]{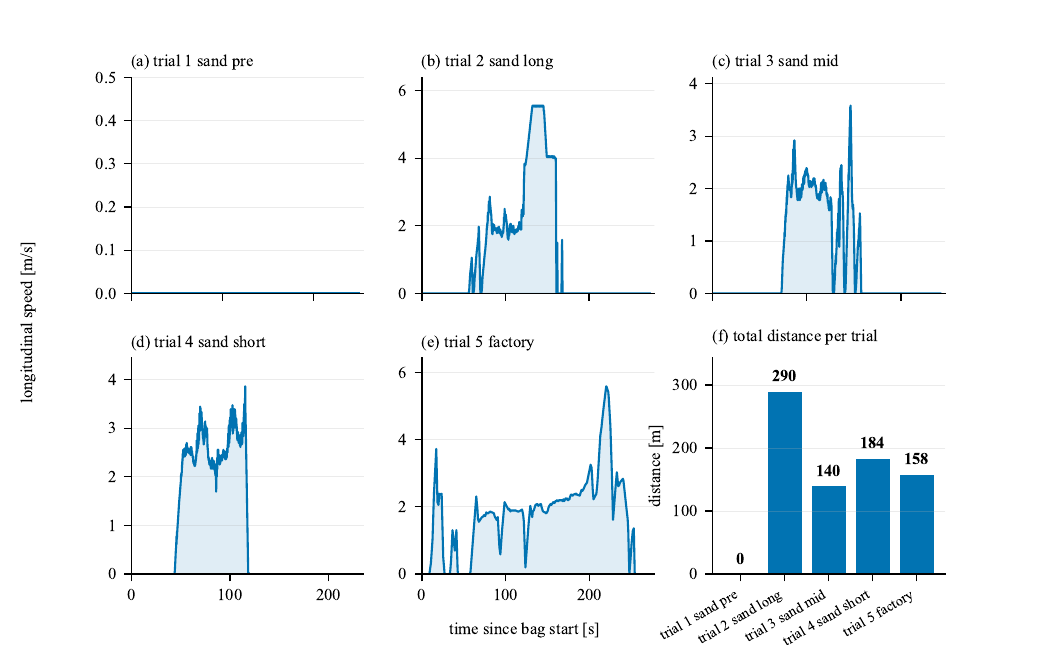}
\caption{Longitudinal velocity measured on Taurus across the five recorded following trials. Trial~1 (\texttt{sand\_road\_real\_track\_1}) is a 12.5\,s stationary pre-roll with zero velocity; the remaining four trials contain motion. All velocities are chassis-feedback values from the ROS bags.}
\label{fig:velocity_timeseries}
\end{figure*}

\begin{figure*}[!t]
\centering
\includegraphics[width=\linewidth]{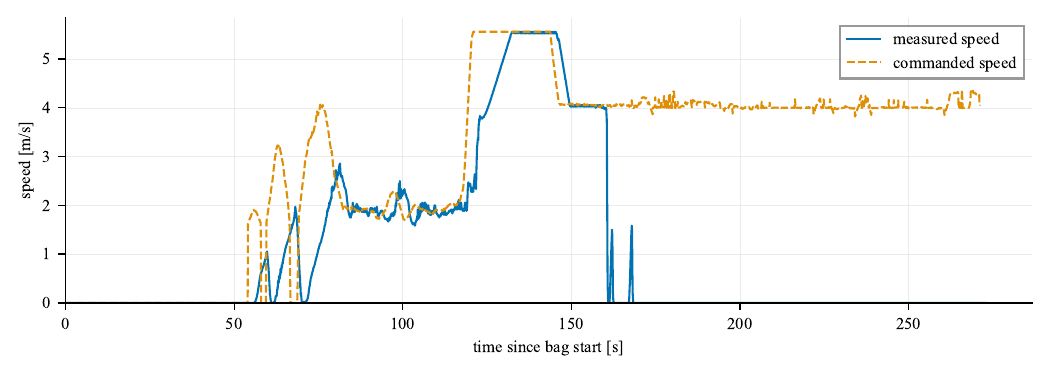}
\caption{Commanded velocity versus measured chassis velocity for trial~2, the longest recorded following run (273.7\,s, 290.4\,m). The figure illustrates the controller's response and lag; it is not a lock-retention or tracking-error metric, because the bags do not record the behavior-layer target state.}
\label{fig:following_response}
\end{figure*}

\begin{figure*}[!t]
\centering
\includegraphics[width=\linewidth]{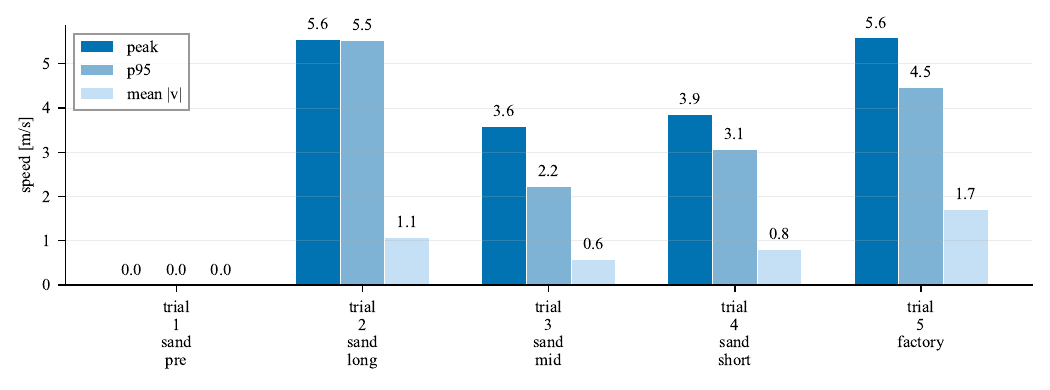}
\caption{Peak, p95, and mean absolute speeds for each Taurus trial. Trial~1 is omitted because it recorded no motion. The behavior-layer \texttt{max\_speed} was configured to 5.0\,m/s; the peak observed speed (5.58\,m/s in trial~5) slightly exceeds this configured limit because of CAN-feedback quantization on a downhill segment. Commanded velocity never exceeded 5.56\,m/s, which remains below the Taurus chassis rated maximum of 9.72\,m/s (35\,km/h).}
\label{fig:speed_summary}
\end{figure*}

\Cref{fig:velocity_timeseries} shows the measured longitudinal velocity for the five Taurus trials, and \cref{fig:following_response} shows command-vs.-measured velocity for the longest run. \Cref{fig:speed_summary} summarizes peak, p95, and mean speeds per trial. All velocity values are chassis feedback from \texttt{/chassis/report}, not GPS; distances are the integral of that velocity over the bag duration.

\begin{figure*}[!t]
\centering
\includegraphics[width=\linewidth]{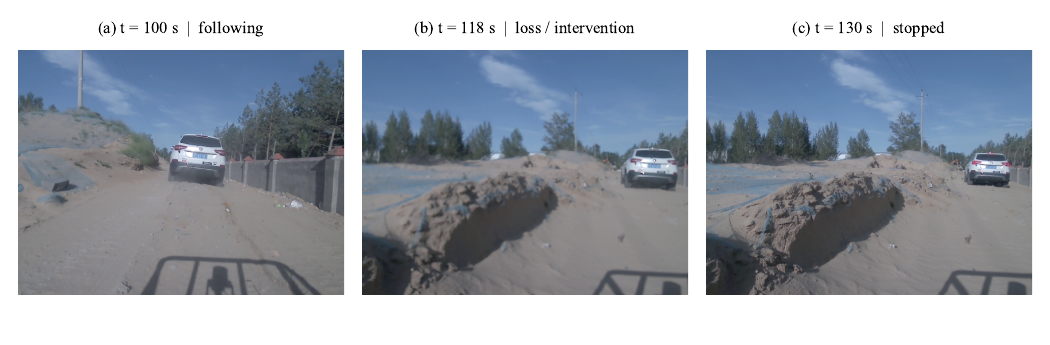}
\caption{Onboard RGB frames from Taurus trial~4 (640$\times$480, YUV422). (a) $t=100$\,s: following the lead vehicle. (b) $t=118$\,s: the operator-labelled loss moment---the lead vehicle is off-center to the front-right. (c) $t=130$\,s: robot stopped after human intervention. The frames are shown only as qualitative evidence of the failure event; they do not support quantitative claims about perception accuracy.}
\label{fig:trial4_camera}
\end{figure*}

\Cref{tab:trials} reports descriptive motion metrics for each bag. The four motion trials covered 771.9\,m in 1,024.92\,s of recorded motion. Trial~4 was explicitly labelled as failed by the operator: the robot followed the lead vehicle from 44.0\,s to 118.6\,s, then lost it off-center to the front-right, received human intervention, and remained stopped from 118.6\,s to the end of the 230.9\,s bag. \Cref{fig:trial4_camera} shows three RGB frames from the onboard camera at 100\,s (following), 118\,s (loss moment), and 130\,s (stopped). The remaining running trials were not individually labelled as success or failure, so we do not aggregate them into a success rate.

\begin{table*}[!t]
\centering
\caption{Descriptive motion metrics for the five recorded Taurus following trials. Distances are integrated from chassis velocity; outcomes are operator labels. These metrics are descriptive and do not by themselves attribute performance to the Silicon Language framework.}
\label{tab:trials}
\begin{tabular}{@{}lrrrrrrp{2.6cm}@{}}
\toprule
Trial & Duration (s) & Distance (m) & Peak (m/s) & Mean (m/s) & p95 (m/s) & Brakes & Outcome \\
\midrule
1 sand\_pre & 12.50 & 0.0 & 0.0 & 0.0 & 0.0 & 0 & stationary pre-test \\
2 sand\_long & 273.71 & 290.43 & 5.556 & 1.074 & 5.528 & 5 & not labelled \\
3 sand\_mid & 242.03 & 140.00 & 3.583 & 0.585 & 2.222 & 4 & not labelled \\
4 sand\_short & 230.88 & 184.01 & 3.861 & 0.802 & 3.056 & 1 & failed (lost lead) \\
5 factory & 278.30 & 157.51 & 5.583 & 1.710 & 4.472 & 4 & not labelled \\
\bottomrule
\end{tabular}
\end{table*}

\textbf{What we do not measure.} The Autoware-era bags record chassis state and control commands but not the behavior-layer topics (\texttt{/behavior/vehicle\_track/*}). Consequently the following higher-level metrics are \emph{not recorded} and must not be claimed: lock-retention rate, safe-distance violation rate, re-acquisition time, and false-positive target rate. Reporting them requires a rerun with the modern \texttt{autodrive\_msgs} schema or a manual replay with the behavior node in the loop; both are scheduled as follow-up work.

\subsection{The Di-Ting downhill failure as a negative gene}

\begin{figure*}[!t]
\centering
\includegraphics[width=\linewidth]{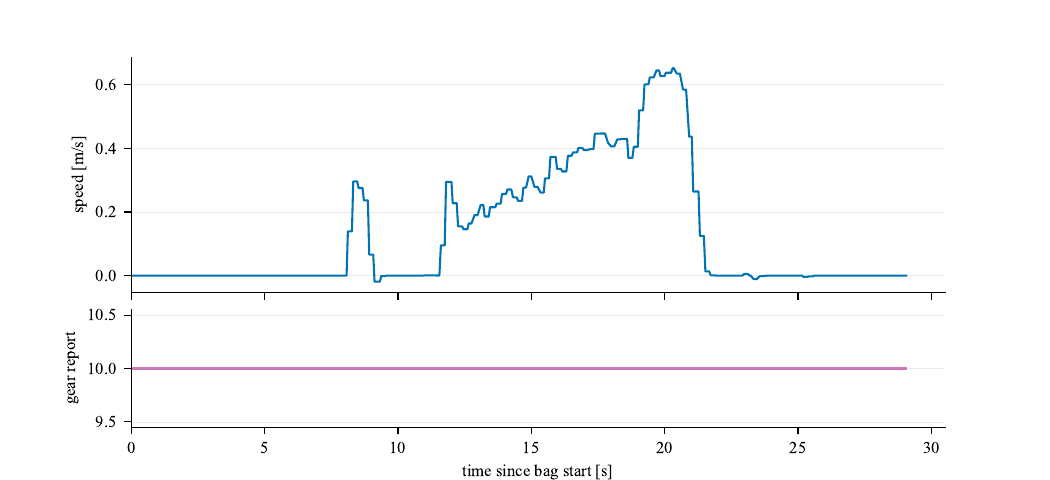}
\caption{Di-Ting velocity profile during the downhill human-track failure. After an initial peak near 0.65\,m/s, the chassis speed collapses to essentially zero as the behavior enters SAFE\_STOP; the gear report remains in drive (value 10), confirming that the stop is commanded by the behavior rather than by a gear shift.}
\label{fig:diting_downhill}
\end{figure*}

The second supporting case is a failure rather than a success. \Cref{fig:diting_downhill} shows the velocity profile from the Di-Ting \texttt{downhill\_lost\_bag}: while descending a slope, the human-track behavior lost the guider because the ground-plane estimate tilted with pitch and clipped the lower body of the person, producing a partial OBB that failed the human-size gate. The state machine transitioned through REACQUIRE to SAFE\_STOP, bringing the robot to a standstill. We preserved this failure as a negative gene, \texttt{kp.meta.negative.downhill\_lost.v1}, so that a receiving robot can avoid the pattern instead of rediscovering it. The pre-screening rule derived from this gene is: if the target platform's typical operating slope exceeds the observed trigger threshold ($\approx$8$^\circ$), do not copy the original ground-removal height threshold blindly.

\subsection{SAFE\_STOP and rollback paths}

\Cref{tab:rollback} lists the two events we document. RB-001 is an observed SAFE\_STOP triggered by the intelligent-following state machine on Di-Ting during its own downhill failure; it is therefore a source-side safety stop, not an adoption rollback after receiving a foreign packet. RB-002 is a coded but unobserved path in the dust-locked following state machine; it is included to show that the safety-stop logic exists, but the paper does not claim it fired in the field. Neither event validates the cross-robot adoption-rollback mechanism described in \cref{subsec:usability}; that mechanism remains part of the framework design and was not exercised in the recorded deployment.

\begin{table*}[!t]
\centering
\caption{Observed SAFE\_STOP events and coded rollback paths. RB-001 is a source-side safety stop, not an adoption rollback; RB-002 is a hypothetical path present in the source code with no corroborating bag.}
\label{tab:rollback}
\begin{tabular}{@{}l>{\raggedright\arraybackslash}p{4.4cm}>{\raggedright\arraybackslash}p{3.6cm}>{\raggedright\arraybackslash}p{4.4cm}>{\raggedright\arraybackslash}p{2.2cm}@{}}
\toprule
ID & Packet & Trigger & Action & Status \\
\midrule
RB-001 & \texttt{kp.behavior.\allowbreak following.\allowbreak intelligent.v1} & Guider lost on downhill; recovery timeout reached & SAFE\_STOP: brakes to $v=0$, publishes \texttt{STATUS\_FAILED} & observed (Di-Ting) \\
RB-002 & \texttt{kp.behavior.\allowbreak following.\allowbreak dust\_locked.v1} & \texttt{miss\_counter} $\geq$ threshold during dust burst & SAFE\_STOP path in state machine & coded, not observed \\
\bottomrule
\end{tabular}
\end{table*}

\subsection{Development time and code scale}

\begin{figure*}[!t]
\centering
\includegraphics[width=\linewidth]{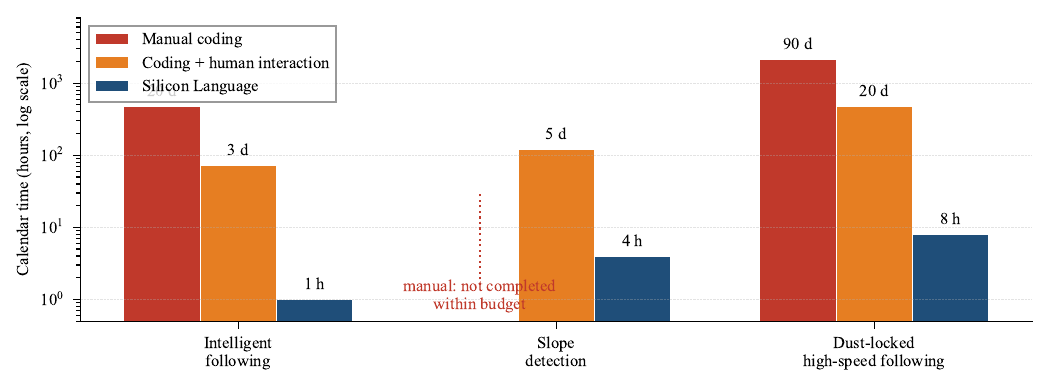}
\caption{Development calendar-time records for three capabilities (log scale). Slope detection by manual coding on the transport robot was not completed within the project budget and is marked as such rather than plotted with a proxy value. Times are calendar-time estimates from project records, not controlled measurements; modes are not like-for-like, so we report no aggregate speedup ratio.}
\label{fig:efficiency}
\end{figure*}

\Cref{fig:efficiency} reports the calendar times recorded in product-release and engineering records for producing the three capabilities under three development modes. We deliberately report the raw recorded times \emph{without} an aggregate speedup figure. The modes are not like-for-like: manual and human+LLM modes include from-scratch development, while the Silicon Language mode records adaptation of an already-existing capability. Quantifying any framework-specific acceleration requires the controlled no-packet and direct-reuse comparisons that are the top priority for follow-up work.

The deployed codebase contains 75,560 shipped lines across 379 files: 64,635 lines of hand-written C++/Python, 3,241 lines of Simulink Embedded Coder generated C, 1,835 lines of ROS \texttt{.msg} definitions, and 5,849 lines of XML/YAML configuration (\texttt{code\_statistics/totals.json}). Third-party drivers (Livox, RSLiDAR, RapidJSON) are excluded. The Simulink-ERT output is deployed code and is counted, but its source models are listed separately. The previously cited ``300k lines'' figure is not supported by this audit and has been removed.

\subsection{Limitations and scope}
\label{sec:discussion}

This deployment is best interpreted as a \emph{proof of concept} for cross-embodiment knowledge exchange. Specific limitations: (1) two robots, one simulated mine environment, one 30-day period; (2) the cross-robot transfers were operator-assisted file copies, not runtime STP exchanges; (3) no controlled with/without-packet or direct-reuse-vs.-adapted comparisons---these are needed to attribute the observed motion to the framework; (4) the recorded metrics are descriptive motion quantities, not the behavior-layer metrics (lock retention, safe-distance compliance, etc.)
needed to validate the following behavior itself; (5) development-time records are descriptive and not like-for-like across modes; (6) fitness weights are heuristic, and cross-robot comparability of normalized review scores is unestablished; (7) multi-reviewer aggregation, branch recombination, and long-horizon fitness dynamics were not exercised---fleet-level evolution remains unvalidated; (8) out-of-distribution generalization and Byzantine robustness in fully decentralized P2P mode are unanalyzed; (9) the 4-D mmWave radar was hardware-installed but not integrated into the recorded trials, so no radar-specific performance claims are made.

\subsection{Code and data availability}
\label{subsec:availability}

The source code, raw ROS bags, and detailed deployment logs used in this study are not publicly released. They contain proprietary implementation details, site-specific information about the coal-mine test facility, and operator-action records that are subject to confidentiality and safety-review obligations. A public specification of the knowledge-packet format and example STP message schemas is planned for future release; until then, qualified researchers may contact the corresponding author for limited, agreement-bound access to anonymized excerpts.

\section{Conclusion}
\label{sec:conclusion}

We presented Silicon Language, a robot-native knowledge exchange framework whose central design principle is that the robot is the subject of its own capability evolution. In the framework, heterogeneous robots encode, publish, retrieve, translate, and review knowledge for one another: a robot that wants a capability searches the hub, translates what it finds to its own sensors and actuators, and judges through its own trial whether the packet is useful and safe. This shift from a human-driven transfer process to a robot-driven exchange process is what makes the analogy to a scholarly system meaningful: the robot, not the human engineer, is the agent that writes, reviews, and adopts knowledge. The framework combines a three-layer infrastructure (edge agent, STP, knowledge hub) with a knowledge-packet abstraction and a receiver-side usability evaluation protocol, organized as a capability stack inspired by the human scholarly system. Progressive blending and automatic rollback are designed to mitigate negative-transfer risk when a robot adopts an external packet.

A 30-day proof-of-concept deployment at an above-ground simulated-mine laboratory in Yulin provided initial evidence: two heterogeneous robots encoded and translated three capabilities across embodiments through operator-assisted file copies mediated by the Silicon Language translation layer. The reported source-side Taurus following trials were recorded on an outdoor sand road and an indoor factory floor, not inside the simulated mine. Per-capability adaptation time dropped from days to hours in our (descriptive, non-controlled) engineering records. The deployed codebase audited for this paper totals 75,560 shipped lines across 379 files.

What the deployment does \emph{not} yet show is fleet-level autonomous evolution: the cross-robot transfers were operator-assisted rather than runtime STP exchanges; multi-reviewer aggregation, branch recombination, and long-horizon fitness dynamics are unvalidated at scale; and the observed motion metrics have not yet been attributed to the framework through controlled with/without-packet comparisons. Priority future work: (1) controlled no-packet and direct-reuse comparisons on the same platforms; (2) a real Di-Ting field trial of the translated dust-locked following packet and a real Taurus field trial of the translated intelligent-following packet; (3) recording behavior-layer topics so that lock-retention, safe-distance, and re-acquisition metrics can be measured; (4) a formal analysis of the acceptance/rollback rule under explicit monitoring assumptions; (5) larger fleets, multiple sites, and longer durations. We will release a public specification of the knowledge packet and example STP traces, together with anonymized evaluation records where confidentiality agreements permit.

\bibliographystyle{IEEEtran}
\bibliography{references}

\end{document}